# Land Cover Mapping Using Ensemble Feature Selection Methods

Gidudu, A.[*§], Abe, B.[§] and Marwala, T.[§]
[§]School of Electrical and Information Engineering
University of the Witwatersrand, 2050, South Africa

**Abstract**
Ensemble classification is an emerging approach to land cover mapping whereby the final classification output is a result of a 'consensus' of classifiers. Intuitively, an ensemble system should consist of base classifiers which are diverse i.e. classifiers whose decision boundaries err differently. In this paper ensemble feature selection is used to impose diversity in ensembles. The features of the constituent base classifiers for each ensemble were created through an exhaustive search algorithm using different separability indices. For each ensemble, the classification accuracy was derived as well as a diversity measure purported to give a measure of the in-ensemble diversity. The correlation between ensemble classification accuracy and diversity measure was determined to establish the interplay between the two variables. From the findings of this paper, diversity measures as currently formulated do not provide an adequate means upon which to constitute ensembles for land cover mapping.

## 1.0 Introduction

Increasingly Earth observation has become a prime source of data in the geosciences and many related disciplines permitting research into the distant past, the present and into the future (Kramer, 2002). This has resulted into new clarity and better awareness of the earth's dynamic nature. Earth observation is based on the premise that information is available from the electromagnetic energy field arising from the earth's surface (or atmosphere or both) and in particular from the spatial, spectral and temporal variations in that field (Kramer, 2002). One of the areas of research interest has always been how to relate Earth observation output e.g. aerial photographs and satellite images to known features (e.g. land cover). The leap from manual aerial photographic interpretation to 'automatic' classification was inspired by the availability of experimental data in various bands in the mid 1960's as a prelude to the launch of the Earth Resources Technology Satellite (ERTS – which was later renamed Landsat 1). This necessitated the adoption of digital multivariate statistical methods for the extraction of land cover information (Landgrebe, 1997). Some of the earliest image classifiers at the time included maximum likelihood and minimum distance to means classifiers (Landgrebe, 2005; Wacker and Landgrebe, 1972). Artificial Neural Network analysis was popular at the time, however the then computational capacity inhibited its wide spread use (Landgrebe, 1997). To date, image classification has benefitted from advancements in improved computational power and algorithm development. An example of the subsequent algorithms that have taken root in image classification include k-Nearest Neighbours, Support Vector Machines, Self Organising Maps, Neural Networks, k-means clustering and object oriented classification.

In light of the improved computational power, variety of classification algorithms, datasets with increasing number of bands, one of the growing areas of interest has been how to 'combine' classifiers in a process better known as ensemble classification. Ensemble classification is premised on combining the outputs of a given number of classifiers in order to derive an accurate classification (Foody et al., 2007). In fields like computational intelligence, combining classifiers is now an established research area (Kuncheva and Whitaker, 2003) and goes by a variety of names such as multiple classifier systems, mixture of experts, committee of classifiers, and ensemble based systems (Polikar, 2006). One of the main prerequisites in building an ensemble system is ensuring that there is diversity among the base (constituent) classifiers (Yu and Cho, 2006). Diversity in ensemble systems may be ensured through the use of different: training datasets, classifiers, features or training parameters (Polikar, 2006). Previous work relating ensemble classification to land cover mapping has focused on investigating how combining different classifiers impacts on classification accuracy (Foody et al., 2007), how different types of ensembles can be applied to land cover mapping (Pal, 2007) and also enforcing diversity through bagging for land cover mapping (Steele and Patterson, 2001). In this paper, ensemble feature selection is investigated as a means of image classification whereby diversity is enforced through using different features. Another key aspect in this paper will be to establish if there is any correlation between classification accuracy and one of the common diversity measures. The paper is arranged as follows: section 2 gives an overview of ensemble classification and diversity measures, section 3 will present the methodology developed to carry out the research, section 4 will present and discuss the results accruing thereof.

## 2.0 Overview of Ensemble Classification

The main idea behind ensemble classification is that one is interested in taking advantage of various classifiers at their disposal to come up with a 'consensus' result. The challenge at hand involves deciding which classifiers to consider and how to combine their results. From the literature (e.g. Polkar, 2006) it is recommended that the constituent classifiers in the ensemble have different decision boundaries, because if identical there will be no gain in combining the classifiers (Shipp and Kuncheva, 2002). Such a set is considered to be diverse (Polkar, 2006). Diversity in ensemble systems has been more commonly explored by considering different classifiers; training a given classifier on different portions of the data; using a classifier with different parameter specifications and using different features. Two methods which have gained prominence in ensemble classification research include bagging or bootstrap aggregating (Breiman, 1996) and Adaboost or reweighting boosting (Freund and Schapire, 1996) which principally involve training a classifier on different training data.

The focus of this paper is ensemble feature selection which entails ensuring diversity through training a given classifier on different features, which in remote sensing would be the different sensor bands. By varying the feature subsets used to generate the ensemble classifier, diversity is ensured since the base classifiers tend to err in different subspaces of the instance space (Oza and Tumer, 2008; Tsymbal et al. 2005) as illustrated in Figure 1. Some of the techniques used to select features to be used in ensemble systems include genetic algorithms (Opitz, 1999), exhaustive search methods and random selection of feature subsets (Ho, 1998).

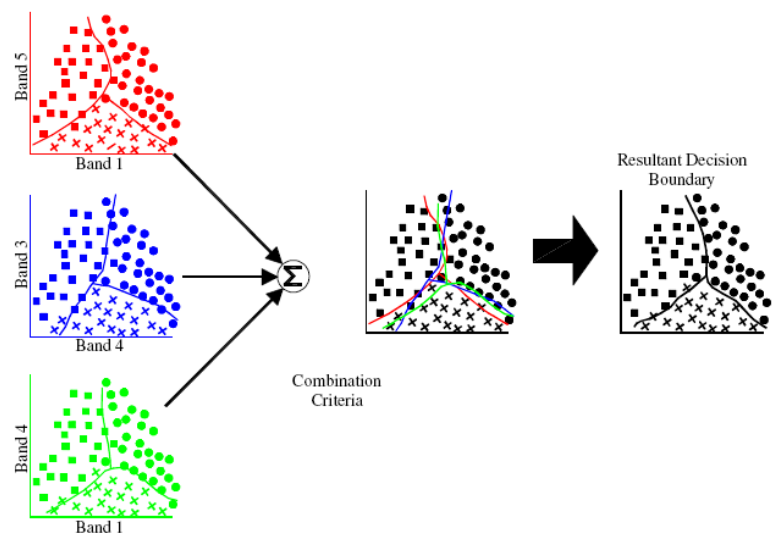

Figure 1: Graphical illustration of an Ensemble classifier system (Adopted from Parikh and Polikar, 2007)

Of equal importance to ensemble classification is how to combine the results of the base classifiers (Foody et al., 2007). A number of approaches exist to combine information from multiple classifiers (Huang and Lees, 2004; Valentini and Masulli, 2003; Giacinto and Roli, 2001) such as majority voting

(Chan and Paelinckx, 2008), weighted majority voting (Polikar, 2006) or more sophisticated methods like consensus theory (Benediksson and Swain, 1992) and stacking (Džeroski and Zenko, 2004).

One of the emerging areas of research interest has been how to quantify diversity, as a result of which numerous diversity measures are under investigation in the literature. The main focus of investigation has centered on finding measures which can be used as a basis upon which to build diverse ensemble systems. In the literature (e.g. Polkar, 2006; Kuncheva and Whitaker 2003), there are two categorizations of diversity measures namely: pair-wise and non-pair-wise diversity measures. Examples of pair-wise measures include: Q statistic, correlation coefficient, agreement measure, disagreement measure and double-fault measure. The diversity measure for the ensemble is derived by calculating the average of the pair-wise measures of the constituent classifiers (Tsymbal et al., 2005; Shipp and Kuncheva, 2002). Non-pair-wise diversity measures include: the entropy measure, Kohavi-Wolpert variance and measurement of inter-rater agreement.

## 3.0 Methodology

The study area for this research was Kampala, the capital of Uganda. The optical bands of a 2001 Landsat image (column 171 and row 60) formed the dataset from which ensembles were created and investigated. There were five land cover classes of interest considered including: water, built up areas, thick swamps, light swamps and other vegetation. Ten ensembles were created each with five base classifiers, the number five having been arbitrarily chosen. For each ensemble, the base classifiers were made up of the bands which yielded the best separability indices (best five band combinations in this case). Three separability indices where used namely: Bhattacharyya distance, divergence and transformed divergence.

For each base classifier and corresponding ensemble and a land cover map was derived using Gaussian Support Vector Machines. The land cover map for each ensemble was consequently derived through majority voting primarily due to its simplicity (Valentini and Masulli, 2002). Each of the derived land cover maps was compared with ground truth data to ascertain their classification accuracy. In order to determine the diversity of each ensemble the kappa analysis was used to give the measure of agreement between the constituent base maps and ultimately the overall ensemble diversity. The influence of diversity on land cover classification accuracy for each ensemble was evaluated by comparing the derived land cover classification accuracies with the derived diversity measures.

## 4.0 Results, Discussion and Conclusions

Table 1 gives a summary of the results depicting the ensembles constituted depending on the separability index used, the respective base classifier classification accuracy assessment and the consequent ensemble classification accuracies.

Table 1: Summary of ensemble classification results

| Index | Bhattacharya Distance | | | | | | Divergence | | | | | | Transformed Divergence | | | | | | None | |
|---|---|---|---|---|---|---|---|---|---|---|---|---|---|---|---|---|---|---|---|---|
| Ensemble | B1 | | B2 | | B3 | | D1 | | D2 | | D3 | | T1 | | T2 | | T3 | | E | |
| | Bands | Acc. | Bands | Acc. | Bands | Acc. | Bands | Acc. | Bands | Acc. | Bands | Acc. | Bands | Acc. | Bands | Acc. | Bands | Acc. | Bands | Acc. |
| | 5,6 | 0.86 | 3,5,6 | 0.87 | 3,5,6 | 0.87 | 2,4 | 0.79 | 2,3,4 | 0.90 | 3,4,5,6 | 0.93 | 2,4 | 0.79 | 1,2,4 | 0.89 | 1,2,5,6 | 0.87 | 1,2,3,4,5 | 0.93 |
| | 3,5 | 0.54 | 1,2,3 | 0.38 | 1,2,3 | 0.38 | 1,5 | 0.73 | 1,2,4 | 0.89 | 2,4,5,6 | 0.92 | 3,4 | 0.86 | 2,3,4 | 0.90 | 1,3,5,6 | 0.87 | 1,2,4,5,6 | 0.93 |
| | 2,3 | 0.90 | 4,5,6 | 0.92 | 4,5,6 | 0.92 | 1,4 | 0.78 | 4,5,6 | 0.92 | 1,4,5,6 | 0.92 | 4,6 | 0.90 | 4,5,6 | 0.92 | 2,3,5,6 | 0.88 | 1,3,4,5,6 | 0.88 |
| | 4,5 | 0.22 | 3,4,5 | 0.92 | 3,4,5 | 0.92 | 3,4 | 0.88 | 1,3,4 | 0.86 | 1,3,5,6 | 0.87 | 1,4 | 0.78 | 2,4,6 | 0.92 | 3,4,5,6 | 0.93 | 2,3,4,5,6 | 0.93 |
| | 1,3 | 0.89 | 2,3,5 | 0.86 | 2,3,5 | 0.86 | 2,5 | 0.76 | 2,4,5 | 0.92 | 2,3,5,6 | 0.88 | 5,6 | 0.86 | 1,3,4 | 0.86 | 1,4,5,6 | 0.92 | 1,2,3,4,6 | 0.93 |
| Accuracy | 0.89 | | 0.89 | | 0.89 | | 0.86 | | 0.90 | | 0.92 | | 0.87 | | 0.90 | | 0.90 | | 0.93 | |
| Diversity | 0.42 | | 0.57 | | 0.70 | | 0.47 | | 0.70 | | 0.72 | | 0.44 | | 0.71 | | 0.65 | | 0.79 | |
| Variance | 0.03 | | 0.02 | | 0.01 | | 0.05 | | 0.01 | | 0.03 | | 0.04 | | 0.01 | | 0.03 | | 0.02 | |

Where Acc. – Accuracy; Diversity – Diversity measure (Agreement); Variance – Diversity Measure (Variance)

It also gives the diversity measure per ensemble according to degree of agreement and variance. From Table 1 it can be observed that for all ensembles, whereas the ensemble classification accuracy was better than many of the base classifiers, in no case was it better than the best classifier within the ensemble. It is, however, critical to note, and the possibility is indicated here and reported elsewhere (e.g. Bruzzone and Cossu, 2004), that whereas the ensemble classification may not be more accurate than all of the base classifiers used in its construction (Foody et al., 2007), it certainly reduces the risk of making a particularly poor selection (Polikar, 2006). Table 1 also shows that across all ensembles, the respective classification accuracy increased as the size of the base classifiers increased. This is further confirmed from Table 2 depicting the binomial tests of significance of the between ensemble classification accuracies. In the simple case of determining if there is a difference between two classifications (2 sided test), the null hypothesis ($H_o$) that there is no significant difference will be rejected if $|Z| > 1.96$ (Congalton and Green, 1998). For each separability index used, increasing the number of features in the base classifiers in general significantly increased the ensemble classification accuracy. The ensemble (E) with five features per base classifier was seen to be significantly better than all the other ensembles apart from D3, where the difference was deemed insignificant. From the results, nothing conclusive can be deduced regarding which of the used separability indices is best suited as a basis upon which to build ensembles.

Table 2: Binomial Test of Significance of between ensemble classification accuracies

|    | B1   | B2   | B3    | D1    | D2   | D3    | T1    | T2   | T3   | E |
|----|------|------|-------|-------|------|-------|-------|------|------|---|
| B1 | -    |      |       |       |      |       |       |      |      |   |
| B2 | 0.44 | -    |       |       |      |       |       |      |      |   |
| B3 | 6.06 | 5.62 | -     |       |      |       |       |      |      |   |
| D1 | 8.20 | 8.64 | 14.22 | -     |      |       |       |      |      |   |
| D2 | 2.09 | 1.65 | 3.97  | 10.28 | -    |       |       |      |      |   |
| D3 | 8.17 | 7.74 | 2.12  | 16.31 | 6.09 | -     |       |      |      |   |
| T1 | 5.58 | 6.02 | 11.61 | 2.63  | 7.67 | 13.71 | -     |      |      |   |
| T2 | 1.47 | 1.03 | 4.59  | 9.67  | 0.62 | 6.71  | 7.05  | -    |      |   |
| T3 | 3.15 | 2.71 | 2.91  | 11.34 | 1.06 | 50.3  | 8.72  | 1.68 | -    |   |
| E  | 9.65 | 9.22 | 3.61  | 17.76 | 7.57 | 1.49  | 15.17 | 8.19 | 6.52 | - |

The relationship between ensemble classification accuracy and diversity was investigated by determining the correlation between ensemble classification accuracy and the agreement measure which in this case was the Kappa value. This was computed by averaging the in-ensemble pair-wise kappa values of the base classifiers measured against each other. In order to get a better appreciation on the in-ensemble diversity, the variance was also computed for the computed pair-wise kappa values. Intuitively, the more diverse the ensemble, the lower the agreement between the classifiers and consequently the lower the consequent kappa values. By extension, the more diverse the ensemble, the bigger the variance between the in-ensemble pair-wise kappa values. Figure 2 depicts the interplay between the ensemble accuracy and diversity measure, which in this case is the mean of the in-ensemble measure of agreement computed from the in-ensemble pair-wise kappa values.

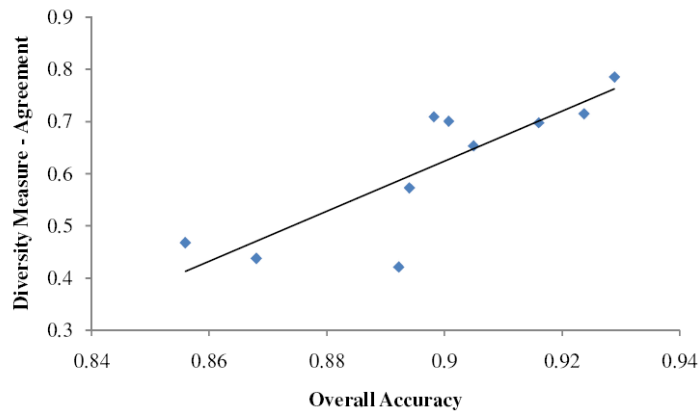

Figure 2: Correlation between Diversity (agreement) and Ensemble classification accuracy

Figure 3 gives a modification of the ensemble measure of agreement whereby instead of considering the mean of the in-ensemble pair-wise kappa values, their variances are considered. The coefficient of correlation of the line of best fit in Figure 2 is 0.83 while in Figure 3 is -0.72. From Figure 2 and 3, respectively, it can be deduced that ensemble accuracy increases as the agreement between the base classifiers increases and as the variance between the base classifier output decreases. In effect, this would ideally imply that the ensemble classification accuracy would increase if there is more agreement between the base classifier outputs. The contradiction this imputes is that to get higher ensemble classification accuracy there is need for less diversity among the base classifiers.

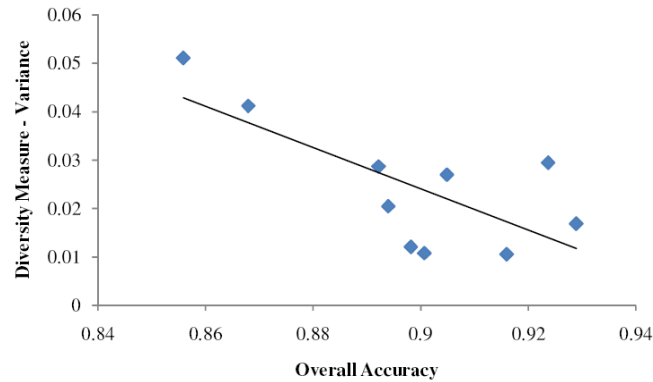

Figure 3: Correlation between Diversity (Variance) and Ensemble classification accuracy

The results bring to the fore the challenge that comes with including diversity measures in ensemble classification research. Clearly its use in determining diversity for land cover mapping is counter intuitive. The problem may stem from using classifier output as the basis upon which to measure diversity. Whereas diversity, as defined in ensemble classification research, is premised on having decision boundaries which err differently, using outputs to determine the measure of diversity presupposes that using different decision boundaries would yield different results. In the case of ensemble feature selection, base classifiers from different features certainly result in decision boundaries which err differently (and hence exhibit diversity), however, their final classification outputs are similar as the high coefficients of correlation depict. Hence basing on the outputs as a measure of diversity clearly gives a poor reflection of how diverse the ensemble is. In their concluding remarks, Shipp and Kuncheva (2002) posit that the quantification of diversity and its use in determining diversity in ensembles will only be possible when a more precise formulation of the notion of diversity is obtained. Until then different heuristics will have to be employed. Whereas ensemble classification presents a unique approach to land cover mapping, the quantification of diversity and its consequent influence in determining the type of ensembles is clearly still open for research.

## Acknowledgements

The authors would like to acknowledge the support of the University of the Witwatersrand, Department of Science and Technology and reviewers.

[*]Corresponding Author: Tel.: +27117177261; Fax: +27114031929
Email: Anthony.Gidudu@wits.ac.za; Anthony.Gidudu@gmail.com